# On interestingness measures of formal concepts

Sergei O. Kuznetsov, Tatiana Makhalova

National Research University Higher School of Economics
Kochnovsky pr. 3, Moscow 125319, Russia
tpmakhalova@hse.ru, skuznetsov@hse.ru

**Abstract.** Formal concepts and closed itemsets proved to be of big importance for knowledge discovery, both as a tool for concise representation of association rules and a tool for clustering and constructing domain taxonomies and ontologies. Exponential explosion makes it difficult to consider the whole concept lattice arising from data, one needs to select most useful and interesting concepts. In this paper interestingness measures of concepts are considered and compared with respect to various aspects, such as efficiency of computation and applicability to noisy data and performing ranking correlation.

## 1 Introduction and Motivation

Formal concepts play an important role in knowledge discovery, since they can be used for concise representation of association rules, clustering and constructing domain taxonomies, see surveys [40, 41, 34]. Most of the difficulties in the application of closed itemsets (or, a more common name, attribute sets) to practical datasets are caused by the exponential number of formal concepts. It complicates both the process of the model construction and the analysis of the results as well. Recently, several approaches to tackle these issues were introduced. For a dataset all closed sets of attributes ordered by inclusion relation make a lattice, below we use the terminology of concept lattices and Formal Concept Analysis [21]. In terms of FCA, a concept is a pair consisting of an extent (a closed set of objects, or transactions in data mining terms) and an intent (the respective closed set of attributes, or itemset).

We propose to divide existing approaches to simplifying concept mining into the groups presented in Figure 1. Methods of the first group compute the concept lattice built on simplified data. The most general way to get smaller lattices is to reduce the size of datasets preserving the most important information. For this purpose such methods as Singular Value Decomposition [15], Non-negative Matrix Decomposition [44] and more computationally efficient k-Means [16], fuzzy k-Means [18], agglomerative clustering of objects based on a similarity function of weighted attributes [17] are used. Another way to reduce the lattice size was proposed in [19]. This approach aims at the reduction in the number of incomparable concepts by making slight changes of the context.

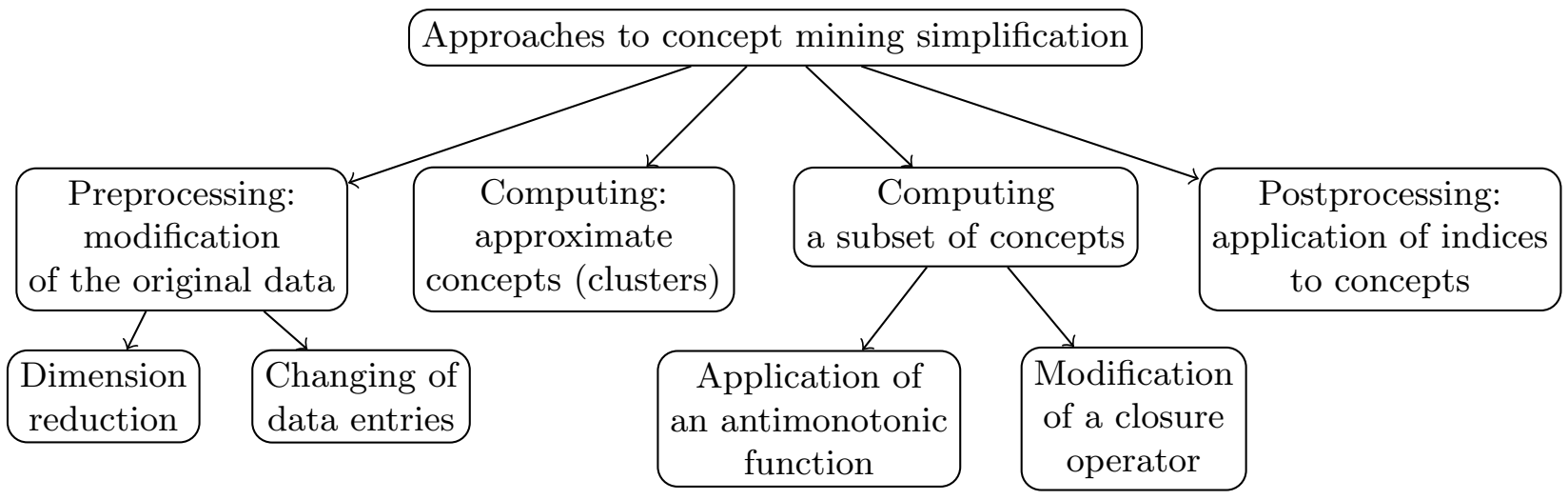

Fig. 1: Classification of methods for concept mining simplification

Computing approximate concepts, or so-called bi-/triclusters, is becoming increasingly more ubiquitous nowadays. In general, biclustering refers to performing simultaneous row–column clustering of real-valued data. Biclusters itself can be defined in different ways [36], e.g. as a submatrix with constant values, with constant values on rows or columns, with coherent values, etc. In the case of binary data bicluster is defined as a submatrix with high density (i.e., the proportion of 1s). In [24] authors presented a set of evaluation criteria for triclusters, which are density, coverage, diversity, noise tolerance and cardinality. As it was shown in experiments, the optimality by some criteria imposes non-optimality w.r.t. other criteria and the optimization by a particular criteria underlies a clustering algorithm itself. Put differently, in the case of multimodal clustering the choice of criteria for cluster evaluation defines the choice of a clustering algorithm rather than selection of a subset of computed clusters.

Methods from "Computing a subset of concepts" class aim at simplifying the analysis of a concept lattice by computing a subset of concepts. It can be done, e.g., by defining an antimonotonic function w.r.t. the size of concept extent. Computing concepts with extents exceeding a threshold was proposed in [35] and studied in relation to frequent itemsets mining in [45], where the authors propose an algorithm for building iceberg lattices. In this case some interesting rare concepts can be discarded, thus, it yields fewer results. From this perspective, the most promising approach is applying $\Delta$-stability. $\Sigma o\varphi\iota\alpha$ algorithm (Sofia) for computing a subset of most $\Delta$-stable concepts was proposed in [10]. $\Delta$-stability will be considered in Section 3.

The most well-reasoned way to build a subset of concepts is to modify the closure operator and to restrict the creation of new concepts by involving background knowledge. In [8] attribute-dependency formulas (AD-formulas) were introduced. The authors define mandatory attributes for particular attributes of a context. If an attribute is included in a closed attribute set without its mandatory attributes the concepts will not be generated. In [5] an analogous approach based on weights of attributes was proposed. The interestingness of concepts is estimated using an aggregation function (average, maximum or minimum), which is applied to a generator, a minimal generator, or to an intent. The authors also note the correspondence between the numerical approach and AD-formulas.

Some restricted closure operators naturally arise from analyzed data and subject area. Carpineto and Romano [13] considered the document–term relation, where objects are documents and terms used in documents are attributes. The authors proposed to modify the closure operator using a hierarchy of terms as follows: two different attributes are considered as equivalent if they have a common ancestor that is a more general term.

In the last decade some polynomial-time algorithms for computing Galois sub-hierarchies were proposed, see [9, 2].

The main idea of methods from the latter group is to assess interestingness of concepts by means of interestingness indices. This approach does not have disadvantages of the previously described methods, namely, getting concepts approximately corresponding to original objects or attributes, missing interesting concepts due to early termination of an algorithm, a costly preprocessing procedure requiring the involvement of domain experts or another reliable source of background knowledge. This class, however, has its own drawbacks, for example, the exponential complexity of the lattice computation can be aggravated by high complexity of index computation. The most promising approach from this point of view is to compute a subset of concepts using the idea of antimonotonicity of an index. When the index value is antimonotonic w.r.t. the order relation of the concept lattice, one can start computing from the top concept and proceed top-down until the concepts with the lowest possible value given by a threshold.

In this paper we focus on a thorough study of concept indices. We leave statistical significance tests of concepts beyond this study, some basic information on statistical approaches to itemsets mining can be found in [48]. The paper is naturally divided into three parts. First, we investigate the main features of indices and discuss the intuition behind them, as well as their applicability in practice with regard to their computation cost. Second, we provide a framework for the development of new indices and give detailed information on the basic metrics and operations that can be used to create new indices. Third, we describe the results of the comparative study of indices regarding the following aspects: estimation of interestingness concepts, approximation of intractable indices and non-sensitivity to noise.

The rest of the paper is organized as follows. Section 2 briefly introduces the main definitions of Formal Concept Analysis. Section 3 is devoted to the description of indices. The main index features are given in Section 3.1. Section 3.2 provides the basic information on indices and respective formulas. In Section 3.3 we use the known indices and their features to build a concept lattice, then we reveal the most interesting concepts (i.e. groups of indices) w.r.t. certain indices. In Section 4 we propose guidelines for the development of new indices based on indices for arbitrary sets of attributes. We discuss the approaches to measuring interestingness and provide the basic metrics and operations that can be used for index construction. Section 5 focuses on a comparative study of the indices w.r.t. the most important tasks: selection of interesting concepts (Section 5.1), approximation of intractable indices (Section 5.2) and noise reduction (Section 5.3). In Section 6 we conclude and discuss the future work.

## 2 Formal Concept Analysis: Basic Definitions

Here we briefly recall the main definitions of FCA [21]. Given a finite set of objects $G$ and a finite set of attributes $M$, we consider an incidence relation $I \subseteq G \times M$ so that $(g, m) \in I$ if object $g \in G$ has attribute $m \in M$. A formal context is a triple $(G, M, I)$. The derivation operators $(\cdot)'$ are defined for $A \subseteq G$ and $B \subseteq M$ as follows:

$$A' = \{m \in M \mid gIm \text{ for all } g \in A\}$$
$$B' = \{g \in G \mid gIm \text{ for all } m \in B\}$$

$A'$ is the set of attributes common to all objects of $A$ and $B'$ is the set of objects sharing all attributes from $B$. The double application of $(\cdot)'$ is a closure operator, i.e., $(\cdot)''$ is extensive, idempotent and monotone. Sets $A \subseteq G$, $B \subseteq M$, such that $A = A''$ and $B = B''$ are said to be closed.

A (formal) concept is a pair $(A, B)$, where $A \subseteq G$, $B \subseteq M$ and $A' = B$, $B' = A$. $A$ is called the (formal) extent and $B$ is called the (formal) intent of the concept $(A, B)$.

A concept lattice (or Galois lattice) is a partially ordered set of concepts, the order $\leqslant$ is defined as follows: $(A, B) \leq (C, D)$ iff $A \subseteq C$ $(D \subseteq B)$, a pair $(A, B)$ is a subconcept of $(C, D)$ and $(C, D)$ is a superconcept of $(A, B)$. Each finite lattice has the highest element with $A = G$, called the top element, and the lowest element with $B = M$, called the bottom element.

| | $a_1$ | $a_2$ | $a_3$ | $a_4$ | $a_5$ | $a_6$ | $a_7$ | $a_8$ | $a_9$ | $a_{10}$ | $a_{11}$ |
|---|---|---|---|---|---|---|---|---|---|---|---|
| stability | × | × | | | | | | | | × | |
| $\Delta_l$ | × | × | × | × | × | × | | × | × | | × |
| $\Delta_h$ (collapse index) | × | × | × | × | × | × | | × | × | | × |
| stab$_{2NOE}$ | × | × | × | × | × | × | | × | × | | |
| stab$_{2OE}$ | × | × | × | × | × | × | | × | × | | |
| stab$_{2OIE}$ | × | × | × | × | × | × | | × | × | | × |
| robustness | × | × | × | × | | × | × | | | × | × |

$a_1$: (designed) for closed subsets, $a_2$: not applicable to arbitrary subsets, $a_3$: based on comparison to other attribute subsets, $a_4$: based on comparison to neighboring attribute subsets, $a_5$: size-based, $a_6$: monotonic–antimonotonic w.r.t. order on attribute sets, $a_7$: using a tuning parameter, $a_8$: polynomial complexity, $a_9$: linear complexity, $a_{10}$: cubic and higher complexity, $a_{11}$: computable in one-pass over data

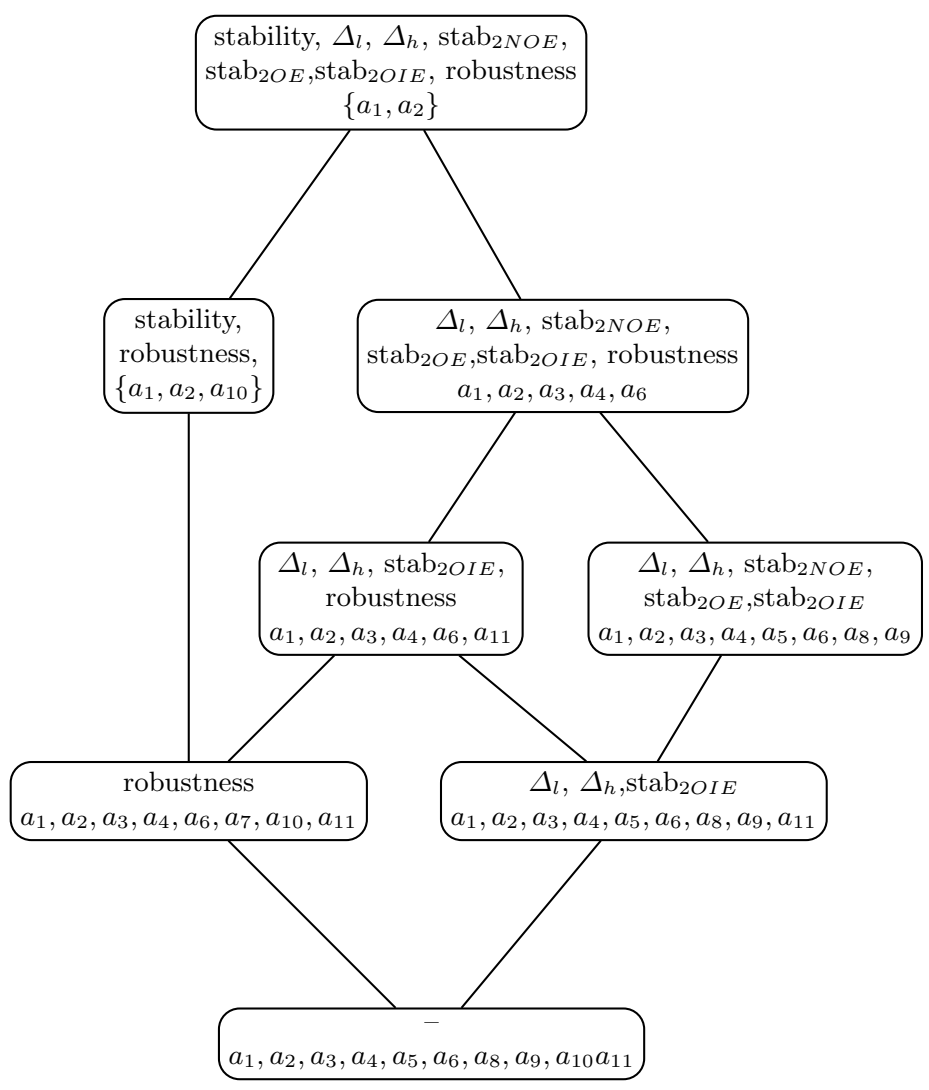

Fig. 2: Formal context of subsampling-based indices (on the left) and the corresponding concept lattice (on the right)

**Example 1** Let us consider a formal context given in Fig. 2. Group of subsampling-based indices for formal concepts is a set of objects, their essential features are attributes. The corresponding concept lattice (see Fig. 2, on the right) consists of 8 concepts.

## 3 Closed Itemset Indices

The application of indices offers advantages over other approaches to reducing of the number of analyzed concepts. For example, it provides a thorough study of a concept lattice without recomputing the lattice and ensures easily interpretable results by preserving objects and attributes corresponding exactly to the original data. However, the exponential time complexity of lattice computation makes attracts attention to antimonotonic indices, since they allow generating only a subset of formal concepts.

In this section we propose a hierarchical clustering of the existing indices for closed set of attributes using the FCA-based approach. We describe the most important index features and use them to build a concept lattice of indices: we consider a formal context of indices and their features, and then compute the ordered set of formal concepts. Upon that we apply some of the described indices to identify the most interesting (top-ranked) concepts.

### 3.1 Description of Index Features

Some of the indices originally designed to assess concepts (*"designed for closed subsets"*) can not be applied to an arbitrary set of attributes (*"not applicable to arbitrary subsets"*), whereas others were originally designed as measures for arbitrary attribute sets. In our study we consider only two indices for arbitrary sets of attributes, i.e. support/frequency, due to their importance in concept mining.

There exist indices which deal with each attribute of an intent (or a whole context) separately and assess the contribution of these attributes to a concept (*"using all of the single attributes of a context"*). Another essential feature of the concept indices is the use of other concepts (*"based on comparison to other attribute subsets"*). The concepts are selected based on a particular condition or index values, i.e., frequent, (dis)similar to some other concepts, etc. In this case an index is comprised of relative values or differences between absolute values of the selected concepts and the assessed concept. We additionally add only one predicate "is neighbor" (*"based on comparison to neighboring attribute subsets"*), since the direct neighbors of a concept are highly important both for computing closed itemsets and for concept mining.

As it was noted previously, support has a special place in concept mining. We define *"support-based (for subsets of attributes)"* and *"support-based (for single attributes)"* features to characterize whether an index is based on support. Support may be regarded as probability of an attribute or a set of attributes, along with it we also consider conditional probability (*"using conditional (joint)* *probability"*). *"Size-based"* indices use the size of an intent or an extent, the total number of attributes or objects, thus, they do not require to assess particular elements of a context.

As it was mentioned before, the antimonotonicity is one of the most important index properties, since it allows one to solve the problem of exponential complexity of algorithms for generating concepts. The time complexity is referred through the following attributes: *"polynomial complexity"*, *"sublinear complexity"*, *"linear complexity"*, *"quadratic complexity"*, *"cubic and higher complexity"*. These groups are defined w.r.t. the size of a context, i.e., $|G||M|$ under the assumption that the lattice has been computed. The possibility to compute indices in *"one-pass over data"* is relevant to complexity problem, but mostly it refers to the memory complexity and analysis of streaming data.

Some indices have a tuning parameter ( *"using a tuning parameter"*) that either changes the concept rankings or is used as a binarization threshold ( *"boolean value"*).

### 3.2 Description of Indices

In our study we use the indices described below.

**Stability** Stability indices for formal concepts were introduced in [30, 31] and modified in [33]. For a formal concept $(A, B)$ the (integral) intentional stability index is the probability that $B$ will remain closed when removing a subset of objects from extent $A$ with equal probability.

The intentional stability is defined as follows:

$$Stab_i\left(A, B\right) = \frac{\mid \{C \subseteq A | C' = B\} \mid}{2^{|A|}}$$

Intentional stability measures overfitting, i.e., the dependence of an intent (i.e.,a closed pattern) on observations (objects of the respective extent). Extensional stability is defined dually.

In [11] it was noted that stability indices are antimonotonic w.r.t. chain of projections. This class of antimonotonicity contains previously known classes [46] and allows one to generate $k$ most stable concepts with delay polynomial in $k$ and the input size.

**Estimates of stability**

*Logarithmic Stability* The problem of computing stability is $\#P$-complete [31], which hampers its application in practice where one usually faces large-sized contexts. In [4] it was proposed to use Monte Carlo approximation of stability, in [12] combination of Monte Carlo and an upper-bound estimate is described. The authors use the logarithmic scale of stability (the concept rankings remain the same):

$$LStab\left(c\right) = -log_2\left(1 - Stab\left(c\right)\right)$$

It allows one to deal with the problem described in [25]: closeness of stability values to 1 for large contexts.

The bounds of stability are given by

$$\Delta_{min}\left(c\right) - log_2\left(|M|\right) \leq -log_2 \sum_{d \in DD(c)} 2^{-\Delta(c,d)} \leq LStab\left(c\right) \leq \Delta_{min}\left(c\right),$$

where $\Delta_{min}\left(c\right) = min_{d \in DD(c)} \Delta\left(c, d\right)$, $DD\left(c\right)$ is the set of all direct descendants of $c$ in a lattice and $\Delta\left(c, d\right)$ is the size of the set-difference between extents of formal concepts $c$ and $d$.

In [32] more precise upper bounds of the logarithmic stability were described. These estimates are based on two direct neighbors. The formulas are given below.

*Max-disjoint-extents upper bound ($stab_{2NOE}$)* The estimate uses two lower neighbors, one of them has the maximal extent among all the lower neighbors, the second one has the maximal extent among the rest of neighbors and does not share elements with the first one:

$$LStab\left(c\right) \leq \Delta_{min}\left(c\right) + \Delta_{min}^{(\emptyset)}\left(c, \Delta_{min}\left(c\right)\right)$$

where $\Delta_{min}\left(c\right) = min_{d \in DD(c)} \Delta\left(c, d\right)$, $\Delta_{min}^{(\emptyset)}\left(c, \Delta_{min}\left(c\right)\right) = d_1 + d_2$, where

$$d_1 = min_{d \in DD(c)} \Delta\left(c, d\right),$$

$$d_2 = min_{d \in DD} \left\{\Delta\left(c, d\right) | d_1 \cap d = \emptyset\right\}.$$

*Max-distinguished-extents upper bound ($stab_{2OE}$)* To compute this index the first neighbor is selected in the same way, while the second one meets the following condition: its extent has the maximal number of objects not included in the extent of the first one.

$$LStab\left(c\right) \leq |d_1| + |d_2| - |d_1 \cap d_2|$$

where $d_2 = min_{d \in DD(c)} \left\{\Delta\left(c, d\right) | d_2 - d_1 = max_{d^* \in DD(c)} d^* - d_1\right\}$ and $d_1 = min_{d \in DD(c)} \Delta\left(c, d\right)$.

*Max-extent upper bound ($stab_{2OIE}$)* The estimate takes into account two different maximal extends:

$$LStab\left(c\right) \leq |d_1| + |d_2| - |d_1 \cap d_2|$$

where $d_1 = min_{d \in DD(c)}\left(c, d\right)$, $d_2 = min_{d \in DD}\left(c\right)$ and $d_1 \neq d_2$. But such a "greedy" strategy can give an underestimated upper bound.

**Stability Indices** The notion of stability was revised and considered with regard to estimating concept-based hypotheses in [31]. Level-wise stability indices are studied. For formal concept $c = (A, B)$ stability index of the $j$th level $(2 \leq j \leq n-1)$ is defined as follows: $J_j(c) = \frac{\gamma_j(c)}{\binom{n}{j}}$, where $n = |A|$, $\gamma_j(c) = |\{Y \subset A \mid |Y| = j, Y' = B\}|$.

Integral stability index is defined as $J_\Sigma(c) = \sum_{i=2}^{n-1} J_i$.

In this study we also consider integral stability indices of the $j$th level $(2 \leq j \leq n-1)$:

– Minor-set-based integral stability: $J_{\Sigma_j}(c) = \sum_{i=2}^{j} J_i$

– Major-set-based integral stability:$J_{\Sigma_j}(c) = \sum_{i=j}^{n-1} J_i$.

Integral stability of the $j$th level may be regarded as the approximation of the stability index. The survey of the problem of best approximation is given in Section 5.

**Robustness** Robustness [46] of an (arbitrary) set of attributes of a dataset has been introduced to estimate the probability that a pattern would still be generated if some transactions (rows) of the dataset are removed. To compute this index one needs to generate $2^{|G|}$ subsamples, where $G$ is the set of transactions (objects). Some classes of itemsets allow computing this index using an exact formula instead of subsampling the data. Here we use the formula for closed sets of attributes. For formal concept $(A, B)$ this is the probability that $B$ will remain closed when removing an object (row) from extent $A$ with probability $1 - \alpha$. Hence, stability index can be considered as an instantiation of robustness for $\alpha = 0.5$. So, for formal concept $c = (A, B)$ the robustness is given as follows:

$$r(c, \alpha) = \sum_{d \leq c} (-1)^{|B_d| - |B_c|} (1 - \alpha)^{|A_c| - |A_d|}$$

**Proposition 1.** *Stability of concept $(A, B)$ is equal to its robustness for $\alpha = 0.5$.*

**Proof**

In [43] it was noted that stability can be computed recursively by the traversal of covering relation (i.e., graph of the diagram) of the concept lattice from the bottom concept upwards as follows:

$$Stab(A, B) = \frac{|\{C \subseteq A \mid C' = B\}|}{2^{|A|}} = \frac{\sigma(A, B)}{2^{|A|}},$$

where $\sigma(A, B) = 2^{|A|} - \sum_{(C,D)<(A,B)} \sigma(C, D)$.

In [4] the following relation of stability to Möbius function $\mu$ [22] was shown:

$$|\{C \subseteq A \mid C' = B\}| = \sum_{(C,D) \leq (A,B)} 2^{|C|} \mu((C, D), (A, B))$$

Using the Möbius function of the concept lattice, the formula takes the following form: $\sigma(A, B) = \sum_{(C,D) \leq (A,B)} 2^{|C|} \mu((C, D), (A, B))$ and stability can alternatively be represented as

$$Stab(A,B) = \sum_{(C,D)\leq(A,B)} 2^{|C|-|A|}\mu\left((C,D),(A,B)\right).$$

Robustness of concept $(A,B)$ is computed as $\sum_{k=0}^{|A|} a_k(1-\alpha)^k$, where

$$a_k = \{\sum_{(C,D)\leq(A,B)} e(D,B) \mid |A|-|C| = k\},$$

and

$$e(D,B) = \begin{cases} 1 & \text{if } D = B \\ -\sum_{(C,D)<(E,F)\leq(A,B)} e(F,B) & \text{otherwise.} \end{cases}$$

This formula gives Möbius function, thus robustness can be rewritten as

$$r((A,B),\alpha) = \sum_{(C,D)\leq(A,B)} \mu\left((C,D),(A,B)\right)(1-\alpha)^{|A|-|C|}.$$

Replacing $1-\alpha$ by 0.5 we obtain the formula for stability, which completes the proof. □

**Concept Probability** In [29] it was noticed that some interesting concepts with the small number of objects (i.e., small extents) usually have low stability values. The concept probability was proposed to get rid of this bias. The definition of concept probability from [29] is equivalent to the concept probability introduced earlier by R. Emilion [20].

The probability that an arbitrary object has all attributes from set $B$ is defined as follows:

$$p_B = \prod_{m\in B} p_m$$

Concept probability is defined as the probability of $B$ being closed:

$$p\left(B = B''\right) = \sum_{k=0}^{n} p\left(|B'| = k, B = B''\right) = \sum_{k=0}^{n}\left[C_k^n p_B^k\left(1-p_B\right)^{n-k}\prod_{m\notin B}\left(1-p_m^k\right)\right]$$

where $n = |G|$.

Concept probability aggregates three probabilistic components: the occurrence of each attribute from $B$ in all $k$ objects, the absence of at least one attribute from $B$ in other objects and the absence of other attributes shared by all $k$ objects.

**Separation** The index was introduced in [29] to estimate the specificity of the object–attribute relation of a concept with respect to the formal context. It is defined as a part of the area covered by a formal concept among all nonzero elements in the rows and columns corresponding to the formal concept.

$$\mathfrak{s}\left(A,B\right)=\frac{|A||B|}{\sum_{g\in A}|g'|+\sum_{m\in B}|m'|-|A||B|}$$

**Frequency (support)** It is one of the most popular measures in the theory of pattern mining. Frequency arises from the assumption that the most "interesting" concepts are frequent ones:

$$supp\left(A,B\right)=\frac{|A|}{|G|}$$

The support provides an efficient level-wise algorithm of semilattice computing since it exhibits antimonotonicity (a priori property [1, 37]):

$$B_1\subset B_2\rightarrow supp\left(B_1\right)\geq supp\left(B_2\right).$$

In this study we say that a set of attributes is frequent if its support exceeds a certain threshold. So, the frequency of an attribute set means that it is frequent.

**Monocle** The Monocle [47] is a method that defines concept weights based on a subset of concepts. The weight function has the following form:

$$w(c,H)=\left(|A|+\sum_{g\in A}N_G(g,H)\right)\cdot\left(|B|+\sum_{m\in B}N_M(m,H)\right),$$

where $N_G(g,H)=|\{(A,B)\mid(A,B)\in H,g\notin A\}|$ is the number of concepts in $H$ not containing an object $g$, $N_M(m,H)$ is defined similarly for attributes, $H\subset\mathcal{L}$. The weight function is monotone w.r.t. the size of $H$.

**$\boldsymbol{\delta}$-Tolerance Closed Frequent Itemsets ($\boldsymbol{\delta}$-TCFIs)** $\delta$-TCFIs [14] uses the subset of lower neighbors (direct descendants) in the concept lattice to assess concept interestingness. The main idea is to select relatively frequent concepts with respect to their direct descendants. Concept $c=(A,B)$ is $\delta$-TCFI iff it meets the following condition: $\forall d=(C,D)$, such that $|D|=|B|+1$, $supp(D)\geq(1-\delta)\cdot supp(B)$, where $\delta\in[0,1]$ is a tolerance factor.

**Margin-closed itemset** The index was proposed by Moerchen et al. [38]. A margin closed itemset has no supersets with almost the same support, by definition, it satisfies the following expression: $X\in FI\;\&\;\forall X'\in FI:X\subset X'\Rightarrow\frac{supp(X')}{supp(X)}\leq1-\alpha$, where $FI$ is a set of frequent itemsets.

As can be seen from the formulas, margin-closed itemset and $\delta$-TCFIs indices are very close to each others. They are based on the same indices computed on different subsets of concepts.

In our study we consider the relaxed formulation of the margin-closed itemset index, namely, we consider the ratio of the maximal extent size among all direct predecessors of a concept to the extent size of the concept itself.

Belohlavek and Trnecka in [6, 7] investigated the group of so-called "basic level" measures. It is a psychology-motivated approach that was designed to formalize the existing psychological approach to defining the basic level of concepts [39]. The group is comprised of similarity- and predictability-based indices, cue validity, category feature collocation and category utility indices.

**Similarity approach (S)** The similarity approach to the assessment of belonging to a basic level was proposed in [42] and subsequently formalized and applied to FCA in [6]. This index is the combination of three fuzzy functions that correspond to formalized properties outlined by E. Rosch [42]: high cohesion of concepts, considerably greater cohesion with respect to upper neighbors and a slightly less cohesion with respect to lower neighbors. The membership degree of the basic level is defined as follows:

$$BL_S = coh^{**}\left(A,B\right) \otimes coh^{**}_{un}\left(A,B\right) \otimes coh^{**}_{ln}\left(A,B\right),$$

where *coh* is a fuzzy function that corresponds to the conditions defined above, $\otimes$ is t-norm [28].

A cohesion function is based on the pairwise similarity of objects from an extent. To assess similarity between two objects authors use simple matching coefficient or Jaccard similarity:

$$sim_{SMC}\left(B_1, B_2\right) = \frac{|B_1 \cap B_2| + |M - \left(B_1 \cup B_2\right)|}{|M|};$$

$$sim_J\left(B_1, B_2\right) = \frac{|B_1 \cap B_2|}{|B_1 \cup B_2|}.$$

A cohesion function is one of the following aggregation functions:

$$coh^{a}_{\ldots}\left(A,B\right) = \frac{\sum_{\{x_1,x_2\}\subseteq A, x_1 \neq x_2} sim_{\ldots}\left(x_1', x_2'\right)}{|A|\left(|A|-1\right)/2}$$

$$coh^{m}_{\ldots}\left(A,B\right) = \min_{x_1,x_2\in A} sim_{\ldots}\left(x_1', x_2'\right)$$

The Rosch's properties for upper and lower neighbors take the following forms:

$$coh^{a*}_{\ldots,un}\left(A,B\right) = 1 - \frac{\sum_{c\in UN(A,B)} coh^{*}_{\ldots}\left(c\right)/coh^{*}_{\ldots}\left(A,B\right)}{|UN\left(A,B\right)|}$$

$$coh^{a*}_{\ldots,ln}\left(A,B\right) = \frac{\sum_{c\in LN(A,B)} coh^{*}_{\ldots}\left(A,B\right)/coh^{*}_{\ldots}\left(c\right)}{|LN\left(A,B\right)|}$$

$$coh^{m*}_{...,un}(A,B) = 1 - \max_{c \in UN(A,B)} coh^{*}_{...}(c) / coh^{*}_{...}(A,B)$$

$$coh^{m*}_{...,ln}(A,B) = \min_{c \in LN(A,B)} coh^{*}_{...}(A,B) / coh^{*}_{...}(c)$$

where $UN(A,B)$ and $LN(A,B)$ are upper and lower neighbors of formal concept $(A,B)$ respectively.

As the authors noted, experiments revealed that the type of a cohesion function does not affect the result, while the choice of a similarity measure can greatly affect the outcome. More than that, in some cases upper (lower) neighbors may have higher (lower) cohesion than the formal concept itself (for example, the boundary cases, when a neighbors's extent / intent is comprised of identical rows / columns). To tackle this issue of non-monotonic neighbors w.r.t. a similarity function the authors proposed to take $coh^{**}_{...,ln}$ and $coh^{**}_{...,un}$ as 0, if the rate of non-monotonic neighbors is larger than a threshold.

Below, we use the following notation: $S^{**}_{SMC}$ and $S^{**}_{J}$, where stars are replaced by the type of cohesion functions for neighbors and objects, respectively. $SMC$ and $J$ stand for simple matching coefficient or Jaccard similarity, respectively.

**Predictability approach (P)** Predictability [7] of a formal concept is computed in a way quite similar to the previous one. In this approach a cohesion function is replaced by the predictability function:

$$P(A,B) = pred^{**}(A,B) \otimes pred^{**}_{un}(A,B) \otimes pred^{**}_{ln}(A,B).$$

From this point of view, concepts are close to basic level if there are only few attributes outside $B$ contained in objects from $A$:

$$E(\mathbb{I}[\langle x,y\rangle \in I] | \mathbb{I}[x \in A]) = -\frac{|A \cap y'|}{|A|} log \frac{|A \cap y'|}{|A|}$$

$$pred(A,B) = 1 - \sum_{y \in M - B} \frac{E(\mathbb{I}[\langle x,y\rangle \in I] | \mathbb{I}[x \in A])}{|M-B|}.$$

**Cue Validity (CV), Category Feature Collocation (CFC), Category Utility (CU)** The following measures are based on the conditional probability of object $g \in A$ given $y \subseteq g'$ [7]:

$$CV(A,B) = \sum_{y \in B} P(A|y') = \sum_{y \in B} \frac{|A|}{|y'|}$$

$$CFC(A,B) = \sum_{y \in M} p(A|y') p(y'|A) = \sum_{y \in M} \frac{|A \cap y'|}{|y'|} \frac{|A \cap y'|}{|A|}$$

$$CU(A,B) = p(A) \sum_{y \in M} \left[ p(y'|A)^2 - p(y')^2 \right] = \frac{|A|}{|G|} \sum_{y \in M} \left[ \left( \frac{|A \cap y'|}{|y'|} \right)^2 - \left( \frac{|y'|}{|G|} \right)^2 \right]$$

CV deals with the probability of an extent given attributes from an intent, CFC takes into account the relation between all attributes from a context and an intent of a formal concept, and CU characterizes how much an attribute from an intent is specific for a given concept rather than for a formal context [49].

### 3.3 Concept Lattice of Indices for Closed Itemsets

In this part of the paper we propose a classification of the existing indices for formal concepts. The classification can be done w.r.t. several features, thus a lot of different index classifications may be suggested. FCA provides a universal framework for discovering domain structure by constructing inclusion-ordered overlapping clusters with various degrees of generality.

We represent the indices described in Section 3.2 and their features from Section 3.1 as a formal context of size $20 \times 19$. The objects and attributes of the formal context are indices and their features, respectively. The context is given in Table 1, the corresponding lattice consists of 73 formal concepts.

We applied the described indices to the lattice to discover interesting concepts and consider top-8 concepts (about 10% of all concepts) by values of the following indices: probability, separation, monocle, margin-closed, frequency, stability, stability estimates, CV, CFC, CU, predictability, similarity approach ($S_J^{aa}$ and $S_{SMC}^{aa}$), robustness with $\alpha = 0.3$. Below, the most frequent concept in the top-ranked groups are listed. We obtained top-ranked concepts with a singleton as an extent or an intent for margin-closed and probability, respectively. The most interesting groups of indices (concepts) with their frequencies (i.e., the rate of top-8 groups where a concept has been included) are listed below.

**Frequency = 0.4**:

Extent: $\Delta_l$, $\Delta_h$, stab$_{2OIE}$, $\delta$-TCFIs.

Intent: for closed subsets, not applicable to arbitrary subsets, based on comparison to other attribute subsets, based on comparison to neighboring attribute subsets, size-based, (anti)monotonic, polynomial complexity, linear complexity, computable in one-pass over data.

Extent: $\Delta_l$, $\Delta_h$ , stab$_{2NOE}$, stab$_{2OE}$, stab$_{2OIE}$, margin-closed*, $\delta$-TCFIs

Intent: for closed subsets, not applicable to arbitrary subsets, based on comparison to other attribute subsets, based on comparison to neighboring attribute subsets, size-based, (anti)monotonic, polynomial complexity, linear complexity.
*The most interesting concepts by stability and its estimates.*

**Frequency = 0.33**:

Extent: $\Delta_l$, $\Delta_h$ , stab$_{2NOE}$, stab$_{2OE}$, stab$_{2OIE}$, similarity, predictability, margin-closed*, $\delta$-TCFIs.

Intent: for closed subsetes, not applicable to arbitrary subsets, based on comparison to other attribute subsets, based on comparison to neighboring attribute subsets, polynomial complexity.
*The most interesting concepts by CV and separation.*

Table 1: A formal context of indices for formal concepts and their key features

| Index name (the number of section) | $a_1$ | $a_2$ | $a_3$ | $a_4$ | $a_5$ | $a_6$ | $a_7$ | $a_8$ | $a_9$ | $a_{10}$ | $a_{11}$ | $a_{12}$ | $a_{13}$ | $a_{14}$ | $a_{15}$ | $a_{16}$ | $a_{17}$ | $a_{18}$ | $a_{19}$ |
|---|---|---|---|---|---|---|---|---|---|---|---|---|---|---|---|---|---|---|---|
| stability (3.2) | × | × | | | | | | | | | × | | | | | | | × | |
| $\mathit{\Delta}_l$ (3.2) | × | × | | | × | × | | | × | | × | | | × | | × | | | × |
| $\mathit{\Delta}_h$ (collapse index)(3.2) | × | × | | | × | × | | | × | | × | | | × | | × | | | × |
| $\text{stab}_{2NOE}$ (3.2) | × | × | | | × | × | | | × | | × | | | × | | × | | | |
| $\text{stab}_{2OE}$ (3.2) | × | × | | | × | × | | | × | | × | | | × | | × | | | |
| $\text{stab}_{2OIE}$ (3.2) | × | × | | | × | × | | | × | | × | | | × | | × | | | × |
| robustness (3.2) | × | × | | | × | | | | | | × | × | | | | | | × | × |
| probability (3.2) | × | × | | × | | | | × | | | | | | × | | × | | | × |
| separation (3.2) | × | | | × | | | | | × | | | | | × | | × | | | |
| support (3.2) | | | × | | | | × | | | | × | | | × | × | | | | × |
| frequency (3.2) | | | × | | | | × | | | | × | × | × | × | × | | | | × |
| monocle (3.2) | × | × | | | | | | | | | | | | | | | | × | |
| $\delta$-TCFIs (3.2) | × | × | | | × | × | × | | × | | × | × | × | × | | × | | | × |
| margin-closed (3.2) | × | × | | | × | | × | | × | | | × | × | | | | | × | × |
| margin-closed* (3.2) | × | × | | | × | × | × | | × | | × | | | × | | × | | | |
| similarity (3.2) | × | × | | | × | × | | | | | | | | × | | | | × | |
| predictability (3.2) | × | × | | × | × | × | | | | × | | | | × | | | | × | |
| CV (3.2) | × | | | | | | | | | × | × | | | × | | × | | | × |
| CFC (3.2) | × | | | × | | | | | | × | × | | | × | | × | | | × |
| CU (3.2) | × | | | × | | | | × | | × | | | | × | | × | | | × |

$a_1$: (designed) for closed subsets, $a_2$: not applicable to arbitrary attribute subsets, $a_3$: applicable to arbitrary attribute subsets, $a_4$: using all of the single attributes of a context, $a_5$: based on comparison to other attribute subsets, $a_6$: based on comparison to neighboring attribute subsets, $a_7$: support-based (for subsets of attributes), $a_8$: support-based (for single attributes), $a_9$: size-based, $a_{10}$: using conditional (joint) probability, $a_{11}$: monotonic–antimonotonic w.r.t. order on attribute sets, $a_{12}$: using a tuning parameter, $a_{13}$: boolean value, $a_{14}$: polynomial complexity, $a_{15}$: sublinear complexity, $a_{16}$: linear complexity, $a_{17}$: quadratic complexity, $a_{18}$: cubic and higher complexity, $a_{19}$: computable in one-pass over data

Extent: $\Delta_l$, $\Delta_h$ , stab$_{2NOE}$, stab$_{2OE}$, stab$_{2OIE}$, separation, margin-closed, margin-closed*, $\delta$-TCFIs.

Intent: for closed subsets, size-based.

*The most interesting concepts by frequency.*

**Frequency = 0.25**:

Extent: $\Delta_l$, $\Delta_h$ , stab$_{2NOE}$, stab$_{2OE}$, stab$_{2OIE}$, robustness, margin-closed*, $\delta$-TCFIs.

Intent: for closed subsets, not applicable to arbitrary subsets, based on comparison to other attribute subsets, based on comparison to neighboring attribute subsets, (anti)monotonic.

Extent: stability, $\Delta_l$, $\Delta_h$ , stab$_{2NOE}$, stab$_{2OE}$, stab$_{2OIE}$, probability, robustness, separation, similarity, predictability, CV, CFC, CU, margin-closed, margin-closed*, monocle, $\delta$-TCFIs.

Intent: for closed.

Extent: $\Delta_l$, $\Delta_h$ , stab$_{2NOE}$, stab$_{2OE}$, stab$_{2OIE}$, robustness, similarity, predictability, margin-closed*, $\delta$-TCFIs.

Intent: for closed subsets, not applicable to arbitrary subsets, based on comparison to other attribute subsets, based on comparison to neighboring attribute subsets.

*The most interesting concepts by monocle.*

Extent: $\Delta_l$, $\Delta_h$ , stab$_{2NOE}$, stab$_{2OE}$, stab$_{2OIE}$, robustness, CV, CFC, margin-closed*, $\delta$-TCFIs.

Intent: for closed subsets, (anti)monotonic.

*The most interesting concepts by CFC.*

Extent: $\Delta_l$, $\Delta_h$ , stab$_{2NOE}$, stab$_{2OE}$, stab$_{2OIE}$, probability, margin-closed*, $\delta$-TCFIs.

Intent: for closed subsets, not applicable to arbitrary subsets, polynomial complexity, linear complexity.

*The most interesting concepts by CU.*

Extent: support, frequency, $\delta$-TCFIs.

Intent: support-based (for attribute sets), (anti)monotonic, polynomial complexity, computable in one-pass over data.

*The most interesting concepts by robustness,* $\alpha = 0.3$.

Extent: CV, CFC.

Intent: for closed subsets, using conditional probability, (anti)monotonic, polynomial complexity, linear complexity, computable in one-pass over data.

*The most interesting concepts by predictability, similarity (Basic Level Metrics).*

As can be seen from the results of the concept rankings, most of the selected indices (accordingly, the rankings) include either $\Delta$-indices or support-based measures. Both types of concepts have almost the same intent that causes the large number of quite similar concepts.

## 4 Generic Approach to the Assessment of Closed Itemsets

In this section we propose guidelines for applying arbitrary itemset indices to closed ones. It should be noted that the problem of the arbitrary itemset assessment was thoroughly investigated in the literature and a lot of different indices for arbitrary sets of attributes and associative rules were proposed. We will not consider in this study statistical-based approaches to patterns and rules assessment (the basic information on the approach can be found in [48]). Some indices for associative rules can be applied directly to evaluate sets of attributes, for example, $lift$ given in Table 2 takes the following form for itemset assessment: $lift(A) = P(A)/\prod_{a \in A} P(a)$. Moreover, a rule-assessment measure may be applied to attribute set within the rule-based approach. In this case, all possible rules are generated from the examined attribute set and then an aggregation function is used to obtain a new index (see details in [48]).

All these indices can be applied directly to formal concepts to assess them either as subsets of objects or as subsets of attributes. However, the closeness of attribute sets can be exploit to construct new indices. The purpose of this section is to provide the main ideas of how to adopt arbitrary itemset indices (including rule-based measures) to assess both dimensions of concepts simultaneously.

To develop a new index one needs to answer the following questions:

- Which concepts are interesting (which patterns we are looking for)?
- Which index will be chosen as the basic measure?
- Which operations will be used to aggregate the values of the basic measure?

The starting point is the personal perspective on concept interestingness. We define the following ones:

- Some "internal"/"external" properties of the concept itself. In this case homogeneous components are considered.
- Impact of the element gathering. Here, a measure is applied to a set and its elements: $f(set)$ and $\{f(element) \mid element \in set\}$.
- Impact of a condition: $f(parameter \mid condition)$ and $f(parameter)$.
- Concept pureness. This approach is based on the comparison of features computed on a concept and out of the concept: $f(parameter)$ and $f(U \setminus parameter))$, where $U$ is the domain of the parameter.
- Stability to random changes in data. Resampling strategy is applied to a set: $\{resampling_\alpha(set)\}$, where $\alpha$ is a noise rate.

The most popular basic measures are given in Table 2 to make the paper self-contained. More detailed information about the listed indices can be found in $[23, 3, 27]$. In the case of formal concepts, indices for associative rules, can be used not only in the ways mentioned above, but also to compare a concept to others similar concepts, i.e., to discover some relatively interesting concepts in a neighborhood.

To combine values of basic measures an aggregation function is used. If a new index measures a relative property of a concepts, then aggregation and comparison operators are applied in any order, i.e., $compare(aggregate_{el \in S}(el), d)$ or $aggregate_{el \in S}(compare(el, d))$, where $S$ is a set of peer elements, $d$ is a distinguished value to compare.

It is worth noting that both real-valued and fuzzy functions can be used to aggregate values. If the values are in the interval $[0, 1]$ (i.e., they can be considered as probabilistic components), a fuzzy aggregation function should be applied. Some binary fuzzy operators ($t$-norm or its dualization $s$-norm related the equation $t(v_1, v_2) = 1 - s(1 - v_1, 1 - v_2)$) and real-valued aggregation functions are given in Table 3. In the case of poor data quality a robust aggregation function, as median, can be applied to data. A comparison function can take one of the following forms: $* - *$, $*/*$, $log(*/*)$.

**Example 2** Let us consider how the proposed guidelines can be used for designing a new index. We will give examples to show that the existing indices conform to the presented approach and to demonstrate how new indices can be constructed step-by-step using this scheme.

*Cue Validity (or CV):*

$$CV\left(A,B\right) = \sum_{y \in B} P\left(A|y'\right) = \sum_{y \in B} \frac{|A|}{|y'|}.$$

- Notion of interestingness: a concept $(A, B)$ is interesting, if the objects from its extent $A$ are more specific for the set of attributes $B$. *We study only a subset of attributes, i.e., properties of the concept itself.*
- Basic measure: conditional probability.
- Aggregation operator (applied to homogeneous elements): sum.

*A new index:*

$$Index_1(A,B) = \min_{(C,D) \in UA((A,B))} \left(PS(D) - PS(B)\right),$$

where $UN((A, B)$ is the set of direct ancestors of $(A, B)$, and $PS(B) = P(B) - \prod_{m \in B} P(m)$.

- Notion of interestingness: a concept $(A, B)$ is interesting if it differs from more general (in the descriptive sense) concepts. *We study only direct ancestors in a lattice.*
- Basic measure: Piatetsky-Shapiro.
- Aggregation operator (applied to non-homogeneous elements): comparison of a concept to each of its ancestors, aggregation by taking the minimal value.

Table 2: Indices for arbitrary itemsets (associative rules $A \to B$)

| **Index** | **Formula** |
|---|---|
| accuracy | $P(AB) + P(\neg A \neg B)$ |
| added value/change of support | $P(B\|A) - P(B)$ |
| certainty factor | $(P(B\|A) - P(B)) / (1 - P(B))$ |
| collective strength | $\frac{P(AB)+P(\neg B\|\neg A)}{P(A)P(B)+P(\neg A)P(\neg B)} \cdot \frac{1-P(A)P(B)P(\neg A)P(\neg B)}{1-P(AB)-P(\neg B\|\neg A)}$ |
| conditional probability | $P(B\|A)$ |
| conviction | $P(A)P(\neg B)/P(A\neg B)$ |
| cosine | $P(AB)/\sqrt{P(A)P(B)}$ |
| Gini index | $P(A)\left(P(B\|A)^2 + P(\neg B\|A)^2\right) +$ <br> $+P(\neg A)\left(P(B\|\neg A)^2 + P(\neg B\|\neg A)^2\right) - P(B)^2 - P(\neg B)^2$ |
| information gain | $\log \frac{P(AB)}{P(A)P(B)}$ |
| J-measure | $P(AB)\log \frac{P(B\|A)}{P(B)} + P(A\neg B)\log \frac{P(\neg B\|A)}{P(\neg B)}$ |
| Jaccard | $P(AB)/\left(P(A) + P(B) - P(AB)\right)$ |
| Klosgen | $\sqrt{P(AB)}(P(B\|A)P(B)),$ <br> $\sqrt{P(AB)}\max(P(B\|A)P(B),\ P(A\|B)P(A))$ |
| Laplace correction | $(N(AB) + 1) / (N(A) + 2)$ |
| least contradiction | $(P(AB) - P(A\neg B)) / P(B)$ |
| leverage | $P(B\|A) - P(A)P(B)$ |
| lift | $P(AB)/\left(P(A)P(B)\right)$ |
| Loevinger | $1 - P(A)P(\neg B)/P(A\neg B)$ |
| normalized mutual information | $\sum_i \sum_j P(A_iB_j) \cdot \log_2 \frac{P(A_iB_j)}{P(A_i)P(B_j)} / \left(-\sum_i P(A_i)\log_2 P(A_i)\right)$ |
| odd multiplier | $(P(AB)P(\neg B)) / (P(B)P(A\neg B))$ |
| Example and <br> counter example Rate | $1 - P(A\neg B)/P(AB)$ |
| odds ratio | $P(AB)P(\neg A\neg B)/\left(P(A\neg B)P(\neg BA)\right)$ |
| one-way support | $P(B\|A)\log_2 \frac{P(AB)}{P(A)P(B)}$ |
| Pearson's $\chi^2$ | $\|G\|\left(\frac{(P(AB)-P(A)P(B))^2}{P(A)P(B)} + \frac{(P(\neg AB)-P(\neg A)P(B))^2}{P(\neg A)P(B)}\right) +$ <br> $+\|G\|\left(\frac{(P(A\neg B)-P(A)P(\neg B))^2}{P(A)P(B)} + \frac{(P(\neg A\neg B)-P(\neg A)P(\neg B))^2}{P(\neg A)P(\neg B)}\right)$ |
| Piatetsky-Shapiro | $P(AB) - P(A)P(B)$ |
| relative risk | $P(B\|A)/P(B\|\neg A)$ |
| Sebag-Schoenaue | $P(AB)/P(A\neg B)$ |
| two-way support | $P(AB)\log_2 \frac{P(AB)}{P(A)P(B)}$ |
| Linear Correlation Coefficient | $\frac{P(AB)-P(A)P(B)}{\sqrt{P(A)P(B)P(\neg A)P(\neg B)}}$ |
| Zhang | $(P(AB) - P(A)P(B)) / \max\left(P(AB)P(\neg B), P(B)P(A\neg B)\right)$ |

Table 3: Aggregation functions

| **Real-valued** | | | |
|---|---|---|---|
| Numerical | | Ordinal (for sorted values $X_{(1)}, X_{(2)}, \ldots, X_{(n)}$) | |
| Sum | $\sum_{v \in V} v$ | Median | $X_{\frac{(n-1)}{2}}$, if $n$ is odd<br>$0.5(X_{\frac{n}{2}} + X_{\frac{n}{2}+1})$,<br>if $n$ is even |
| Arithmetic mean | $\sum_{v \in V} v//\vert V\vert$ | Maximum | $X_{(n)}$ |
| Geometric mean | $\sqrt[\vert V\vert]{\prod_{v \in V} v}$ | Minimum | $X_{(1)}$ |
| Harmonic mean | $\vert V\vert/(\sum_{v \in V} v^{-1})$ | Midrange | $0.5(X_{(1)} + X_{(n)})$ |

| **Fuzzy** | | | |
|---|---|---|---|
| T-norms | | S-norms | |
| Drastic product | $min(v_1, v_2)$,<br>if $max(v_1, v_2) = 1$<br>0, otherwise | Drastic sum | $max(v_1, v_2)$,<br>if $min(v_1, v_2) = 0$<br>1, otherwise |
| Bounded difference | $max(0, v_1 + v_2 - 1)$ | Bounded sum | $min(1, v_1 + v_2)$ |
| Einstein product | $\frac{v_1 v_2}{2-(v_1+v_2-v_1 v_2)}$ | Einstein sum | $\frac{v_1+v_2}{1+v_1 v_2}$ |
| Algebraic product | $v_1 v_2$ | Probabilistic sum | $v_1 + v_2 - v_1 v_2$ |
| Hamacher product | $\frac{v_1 v_2}{v_1+v_2-v_1 v_2}$ | Hamacher sum | $\frac{v_1+v_2-2v_1 v_2}{1-v_1 v_2}$ |
| Minimum | $min(v_1, v_2)$ | Maximum | $max(v_1, v_2)$ |

*A new index:*

$$Index_2(A, B) = |M \setminus B| \sum_{m \in M \setminus B} 1/P(m|B).$$

– Notion of interestingness: a concept $(A, B)$ is interesting if it stands out from the context. *We study out-of-concept elements of a context.*
– Basic measure: conditional probability.
– Aggregation operator (applied to homogeneous elements): harmonic mean.

## 5 Quality of Indices

Application of indices in practice is aimed mostly at addressing the following goals: selecting a subset from the whole set of concepts or computing only a fragment of a concept lattice. The first goal then is attained by selecting the most interesting concepts from the set of concepts or detecting “original” concepts computed on a noisy context. The second goal is to reduce the computation cost by constructing only the most interesting concepts.

### 5.1 Rank Correlation

To realize how similar the indices are, we examined the similarity of the concept rankings by values of indices. The pairwise similarity of indices is measured by the Kendall tau correlation coefficient [26], this coefficient takes into account not only an absolute rank, but also a relative position.

We randomly generated 4 groups of formal concepts with the following densities (the rate of the "1" in the context): 0.1, 0.2, 0.3, 0.4. Each group consisted of 100 formal contexts with the number of attributes ranged between 10 and 50, and the number of objects varying from 40 to 80.

The standard deviation of within-group pairwise correlation is quite small (not more then 0.05) and does not depend on the context density ($p$-values of Levene's test is less than 0.05). Since the studied values cannot be assumed to be normally distributed (based on D'Agostino and Pearson's test), the Wilcoxon's test was used to compare mean values of the Kendall tau coefficient.

The averaged values of the pairwise Kendall tau coefficients are presented in Table 4. Among investigated indices several groups of correlated measures stood out, but only few values of pairwise correlation are statistically stable (have the same average value w.r.t. context densities).

One of the groups of correlated indices corresponds to approximate robustness with different values of parameters. It allows us to conclude that relative importance of concepts is mostly preserving regardless of $\alpha$. Another class of indices that utilizes similarity and predictability approaches (Basic Level Metrics group) yields the second group. It should be noted that the highly correlated indices of this group are the measures based on similarity approach with the same cohesion function and with different aggregation functions for sub/superordinate concepts. The conclusion agrees with the results presented in [6].

The other groups of correlated indices are highlighted in Table 4. Stability (robustness) is the most complex index for calculation, hence it is important to identify more easily computable indices and use them instead. In our experiments we found that logarithmic estimates of stability ($\Delta_l$, $\Delta_h$) and Max-distinguished-extents upper bound ($\text{stab}_{2OE}$) are strongly correlated with stability, highly correlated robustness indices belong to the same complexity class as stability.

### 5.2 Integral Stability Indices: Best Approximation of Stability

Here, we consider the problem of stability approximation w.r.t. different levels of integral stability index, i.e., the number of levels that are used to compute the index. Since the extent size of concepts varies in the range from 0 to $|G|$, for each concept we took the level of stability depending on the size of set $A$, to be precise, the level is defined by formula $[rate * |A|]$, where $0 < rate < 1$.

To find the best $rate$ we used randomly generated formal contexts (described in the previous subsection) of different densities: 0.1, 0.2, 0.3 and applied simple linear regression. The integral stability index of the $j$th level was taken as a regressor and the stability as a dependent variable.

Scatter plots of the studied indices are given in Figure 3. As we can see from the diagrams, taking too small rates does not allow us to estimate stability, since for most concepts small-sized subsets $Y \subset A$, such that $Y' = B$ do not exist.

The first local maximum of the coefficients of determination for model

$$stability = A * stability_{\Sigma_{ratio}} + B$$

Table 4: The averaged Kendall tau coefficient for indices. Statistically equal average values are in bold type.

| | $\Delta_l$ | $\Delta_h$ | $stab_{2NOE}$ | $stab_{2OE}$ | $stab_{2OIE}$ | $rob_{0,1}$ | $rob_{0,3}$ | $rob_{0,5}$ | $rob_{0,8}$ | probability | separation | support | marg-clos* | $S^{aa}_{SMC}$ | $S^{am}_{SMC}$ | $S^{ma}_{SMC}$ | $S^{mm}_{SMC}$ | $S^{aa}_{J}$ | $S^{am}_{J}$ | $S^{ma}_{J}$ | $S^{mm}_{J}$ | P | CV | CFC | CU |
|---|---|---|---|---|---|---|---|---|---|---|---|---|---|---|---|---|---|---|---|---|---|---|---|---|---|
| stability | 0,88 | 0,73 | 0,48 | 0,63 | 0,54 | 0,86 | 0,94 | **1,00** | 0,97 | 0,27 | 0,18 | 0,32 | **0,12** | 0,17 | 0,16 | 0,05 | 0,05 | 0,18 | 0,18 | 0,07 | **0,08** | -0,02 | 0,05 | 0,06 | 0,07 |
| $\Delta_l$ | | 0,55 | 0,34 | 0,50 | 0,40 | 0,88 | **0,93** | 0,88 | 0,83 | 0,15 | 0,13 | 0,18 | **0,20** | 0,16 | 0,16 | **0,00** | **0,00** | **0,13** | **0,13** | **0,01** | **0,02** | -0,06 | 0,00 | -0,02 | -0,05 |
| $\Delta_h$ | | | 0,66 | 0,67 | 0,65 | 0,53 | 0,64 | 0,73 | 0,76 | 0,55 | 0,28 | 0,65 | -0,07 | 0,16 | 0,16 | 0,15 | 0,15 | 0,29 | 0,30 | 0,19 | 0,21 | 0,08 | 0,16 | 0,27 | 0,34 |
| $stab_{2NOE}$ | | | | 0,88 | 0,95 | 0,32 | 0,41 | 0,48 | 0,51 | 0,67 | 0,36 | **0,80** | -0,47 | 0,34 | 0,36 | 0,35 | 0,35 | 0,53 | 0,53 | 0,45 | 0,47 | 0,30 | 0,21 | 0,38 | 0,50 |
| $stab_{2OE}$ | | | | | 0,94 | **0,47** | 0,56 | 0,63 | 0,66 | 0,60 | 0,32 | 0,72 | -0,37 | 0,38 | 0,38 | 0,39 | 0,39 | 0,54 | 0,55 | 0,47 | 0,48 | 0,34 | 0,18 | 0,29 | 0,40 |
| $stab_{2OIE}$ | | | | | | 0,38 | 0,47 | 0,54 | **0,57** | **0,65** | 0,36 | 0,78 | -0,45 | 0,36 | 0,37 | 0,37 | 0,37 | 0,53 | 0,54 | 0,46 | 0,47 | 0,32 | 0,21 | 0,35 | 0,47 |
| $rob_{0.1}$ | | | | | | | 0,91 | 0,86 | 0,81 | 0,14 | 0,12 | 0,17 | 0,22 | 0,17 | 0,16 | 0,02 | 0,02 | 0,12 | 0,13 | 0,02 | **0,03** | -0,04 | **-0,02** | -0,05 | -0,05 |
| $rob_{0.3}$ | | | | | | | | 0,94 | 0,89 | 0,21 | 0,15 | 0,26 | 0,16 | 0,17 | 0,16 | 0,04 | 0,03 | 0,16 | 0,16 | 0,05 | **0,06** | -0,03 | 0,02 | 0,01 | 0,02 |
| $rob_{0.5}$ | | | | | | | | | 0,97 | 0,27 | 0,18 | 0,32 | 0,12 | 0,17 | 0,16 | 0,05 | 0,05 | 0,18 | 0,18 | 0,07 | **0,08** | -0,02 | 0,05 | 0,06 | 0,07 |
| $rob_{0.8}$ | | | | | | | | | | 0,30 | 0,19 | 0,36 | 0,09 | 0,16 | 0,16 | 0,06 | 0,05 | 0,19 | 0,19 | 0,08 | **0,09** | -0,01 | 0,07 | 0,08 | 0,10 |
| probability | | | | | | | | | | | 0,27 | 0,81 | -0,51 | 0,22 | 0,24 | 0,29 | 0,30 | 0,44 | 0,44 | 0,35 | 0,38 | 0,28 | -0,02 | 0,31 | 0,36 |
| separation | | | | | | | | | | | | 0,40 | -0,27 | 0,25 | 0,27 | 0,17 | 0,17 | 0,24 | 0,25 | **0,26** | **0,25** | 0,14 | 0,40 | 0,25 | 0,18 |
| support | | | | | | | | | | | | | **-0,62** | 0,27 | 0,29 | 0,32 | 0,33 | 0,53 | 0,53 | 0,43 | 0,45 | 0,28 | 0,23 | 0,48 | 0,61 |
| marg-clos* | | | | | | | | | | | | | | -0,26 | -0,28 | -0,37 | -0,38 | -0,47 | -0,47 | -0,50 | -0,51 | -0,39 | -0,19 | -0,36 | -0,53 |
| $S^{aa}_{SMC}$ | | | | | | | | | | | | | | | 0,91 | 0,58 | 0,57 | 0,61 | 0,61 | 0,50 | 0,51 | 0,52 | 0,07 | **0,03** | 0,16 |
| $S^{am}_{SMC}$ | | | | | | | | | | | | | | | | 0,57 | 0,57 | 0,62 | 0,63 | 0,50 | 0,50 | 0,51 | 0,08 | 0,05 | 0,17 |
| $S^{ma}_{SMC}$ | | | | | | | | | | | | | | | | | 0,97 | 0,59 | 0,59 | 0,68 | 0,69 | 0,73 | **0,00** | 0,01 | 0,16 |
| $S^{mm}_{SMC}$ | | | | | | | | | | | | | | | | | | 0,59 | 0,59 | 0,67 | 0,68 | 0,73 | **0,00** | 0,01 | 0,16 |
| $S^{aa}_{J}$ | | | | | | | | | | | | | | | | | | | 0,97 | 0,61 | 0,62 | 0,55 | 0,11 | 0,17 | 0,36 |
| $S^{am}_{J}$ | | | | | | | | | | | | | | | | | | | | 0,61 | 0,62 | 0,55 | 0,11 | 0,17 | 0,35 |
| $S^{ma}_{J}$ | | | | | | | | | | | | | | | | | | | | | 0,93 | 0,65 | 0,12 | 0,11 | 0,30 |
| $S^{mm}_{J}$ | | | | | | | | | | | | | | | | | | | | | | 0,66 | 0,11 | 0,12 | 0,29 |
| P | | | | | | | | | | | | | | | | | | | | | | | -0,05 | -0,03 | 0,12 |
| CV | | | | | | | | | | | | | | | | | | | | | | | | 0,35 | 0,39 |
| CFC | | | | | | | | | | | | | | | | | | | | | | | | | 0,42 |

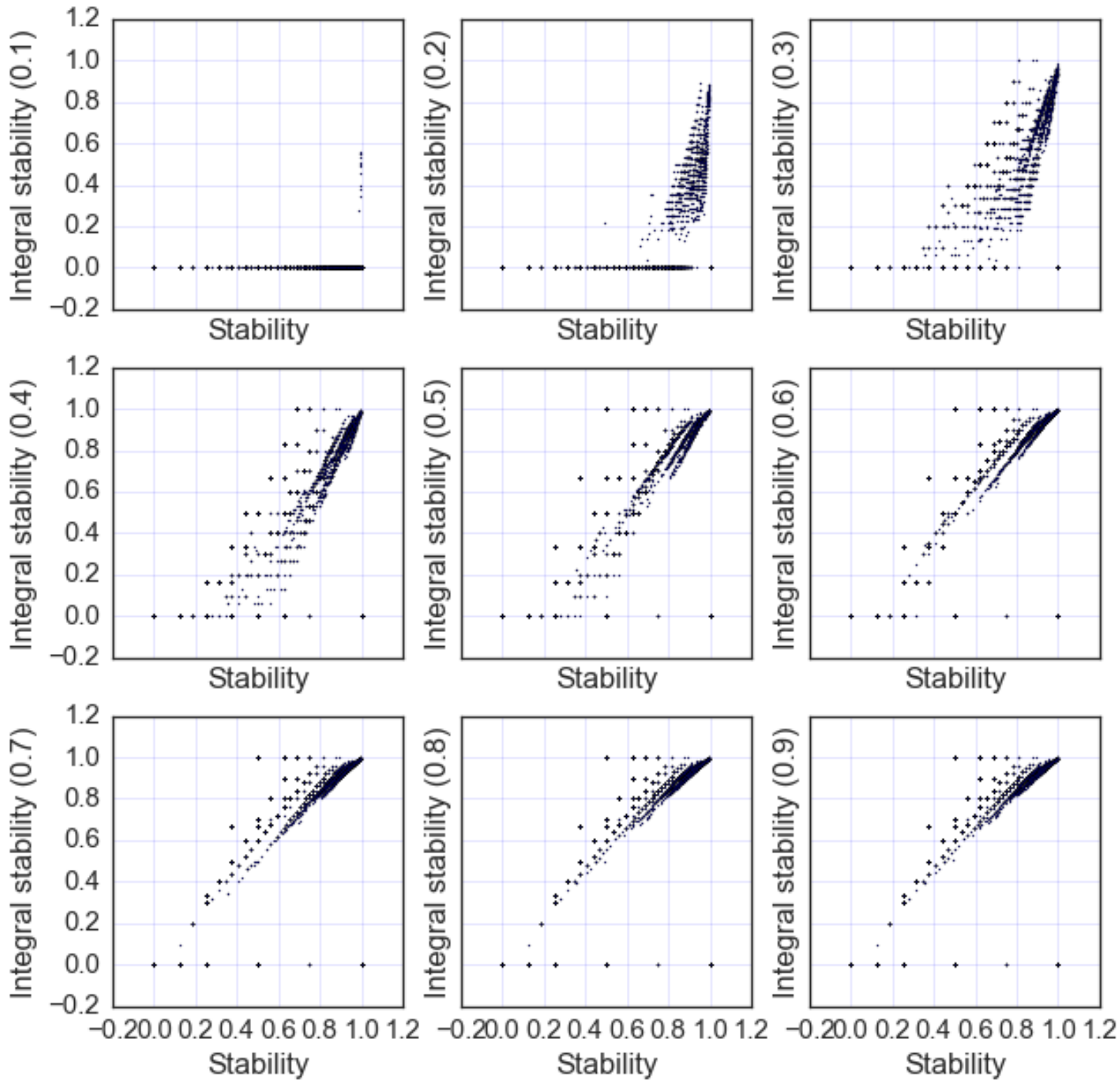

Fig. 3: The dependence of stability values on integral stability values

corresponds to $rate = 0.4$ (0.78 and 0.79 for context densities 0.2 and 0.3, respectively). Thus, this value can be used as the most suitable one for computing estimates of stability by integral stability index.

### 5.3 Noise Filtering

In practice one usually faces noisy data. Even a small noise rate can result in exponential explosion of the number of formal concepts [29]. In this connection, we study the ability of indices to select original concepts from the set of concepts computed on a noisy context. We took 5 formal concepts, the lattices of the first four of them have quite simple structure (see Figure 4) and the last one is a fragment of Mushroom dataset [1] consisted of 500 objects and 14 attributes.

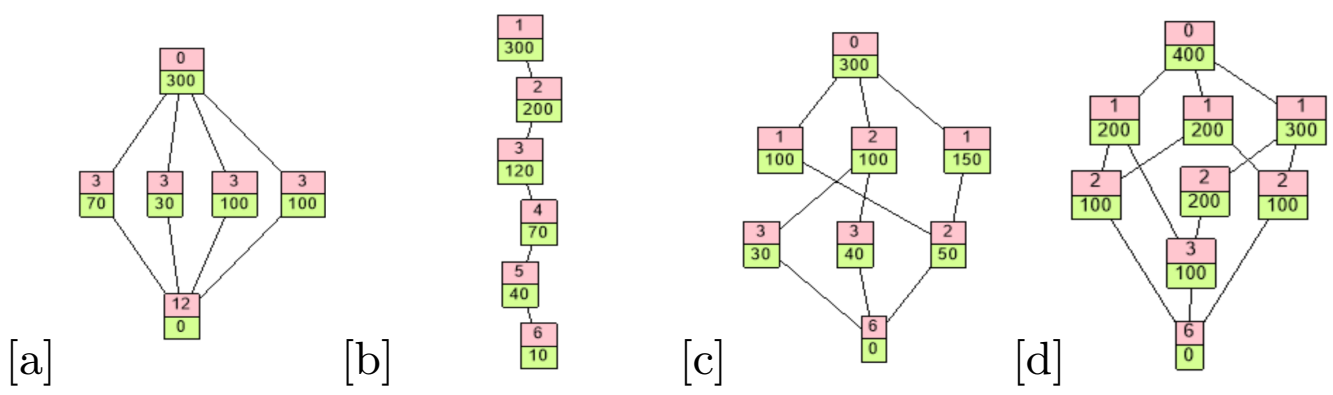

Fig. 4: The concept lattices of formal contexts with 300 objects and 6 attributes (a - c), with 400 objects and 4 attributes (d)

We added different amount of noise into the contexts as follows: each entry changes with given probability (noise rate). For "noisy" lattices we identified original concepts and considered a binary classification problem. We computed the AUC (Area Under the Receiver Operator Curve) from each index separately. The averaged AUCs within groups corresponding the same datasets with different noise rates and within groups with the same noise rate for the described above contexts are given in Figure 5.

As can be seen in Figure 5, the index quality mostly depends on lattice structure rather than on noise rate. Cue Validity (CV), Category Feature Collocation (CFC), Category Utility (CU) and separation have the highest AUC on particular datasets, but the quality considerably changes depending on the lattice structure. For instance, AUC of CFC varies from 0.67 to 0.95. The most stable results correspond to robustness with $\alpha \in \{0.3, 0.5, 0.8\}$, the estimates of stability are able to distinguish most of the original concepts (the AUC is greater than 0.7).

Almost all indices are stable with respect to different noise rates. The poor data quality has the strong impact on the estimates of stability: the quality of these indices drops down as the noise rate increases. Since these measures are based on the elements of a lattice, the more noise is introduced, the more noisy concepts are involved in computation. The quality of similarity-based indices (Basic Level group) is close to random guessing, which makes them inapplicable to the analysis of noisy data.

[1] https://archive.ics.uci.edu/ml/datasets/Mushroom

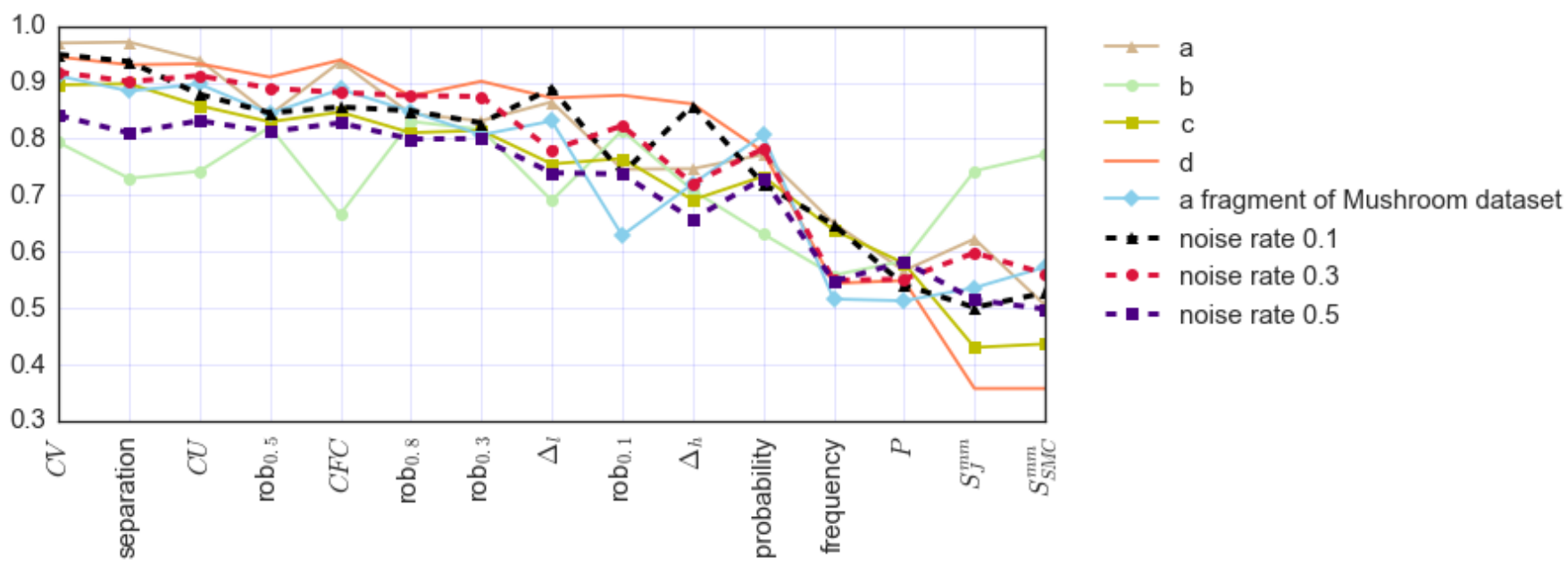

Fig. 5: The averaged AUC within groups with the same noise rate (dotted lines) and corresponding to the same datasets (lines a-d)

The experiments allow us to conclude that the CV, CU, CFC, separations and robustness are the most suitable for the analysis of noisy data.

## 6 Conclusion

In this paper we have presented some results on formal concept indices. We have also analyzed the existing indices for closed itemsets in Data Mining. We defined the main features of the indices and proposed their classification using an FCA-based approach. We have also provided some examples of most interesting groups of indices selected by means of some of the studied indices.

We have given basic ideas of adapting indices for arbitrary itemsets to closed ones. We have suggested to utilize bimodal nature of the concepts to get new indices on the basis of the indices for arbitrary set of attributes.

An important part of the study is devoted to practical application of indices. We have performed a comparative study of indices in the context of the following tasks: selection of the most interesting concepts, approximation of the exponentially computable indices (stability) and filtering noisy data. The results of our experiments allow us to distinguish groups of correlated indices, thus these results can be used to reduce computational complexity of concept mining by choosing easily computable indices among the correlated ones. Another important aspect of our study is identification of indices that can be used for the analysis of noisy data. It was shown that the noise filtering quality of indices depends more on the structure of the concept lattice than on the noise rate. The strongest dependence of the filtering quality on noise rate corresponds to the estimates of stability. Even a small proportion of noise added to data can significantly change the lattice structure, which results in biased values of neighbor-based indices.

As possible directions of future work, we can propose the study of indices for approximate concepts, e.g., biclusters, as well as indices for multimodal concepts. Another important application of indices as tools for selecting hypothesis can be examined within the framework of classification and learning.

## Acknowledgments

This paper was prepared within the framework of the Basic Research Program at the National Research University Higher School of Economics (HSE) and supported within the framework of a subsidy by the Russian Academic Excellence Project '5-100'.

**List of indices**

| **Used notation** | **Description** |
|---|---|
| stability (3.2) | It estimates the probability of the closeness of an attribute set after removing some elements from the object set. |
| $\Delta_l$ (3.2) | $\Delta$-measure that is based on direct descendents of a concept. |
| $\Delta_h$ (3.2) | $\Delta$-measure that is based on the maximal extent. |
| $\text{stab}_{2NOE}$ (max-disjoint-extents upper bound, 3.2) | $\Delta$-measure that is based on two maximal disjoint extents. |
| $\text{stab}_{2OE}$ (max-distinguished-extents upper bound, 3.2) | $\Delta$-measure that is based on the maximal extent and the extent with maximal disjoint subset. |
| $\text{stab}_{2OIE}$ (max-extent upper bound, 3.2) | $\Delta$-measure that is based on two maximal extents. |
| robustness (3.2) | It estimates the probability that an attribute set will be arose from the data, if some transactions are removed. |
| probability (3.2) | It estimates the probability that a set of attributes is closed, if all attributes are independent and the probability of an attribute is assumed to be equal to its frequency. |
| separation (3.2) | The ratio of the area covered by a formal concept to all nonzero elements in rows and columns corresponding to the concept. |
| frequency, support (3.2) | The frequency of an attribute set (if all attributes are assumed to be independent). |
| monocle (3.2) | The index assigns weights to concepts by taking into account the size of a context and the number of the concepts that do not contain objects and attributes contained to the concept. |
| $\delta$-TCFIs (3.2) | It selects relatively frequent concepts with respect to their direct descendants. |
| margin-closed (3.2) | The index characterizes whether a frequent intent have well-distinguishable support w.r.t. other frequent intents, supersets of the intent. |
| $S_*^{**}$ (similarity, 3.2) | A fuzzy combination of three cohesion functions which correspond to the following conditions: high cohesion of concepts, considerably greater cohesion with respect to upper neighbors and a slightly less cohesion with respect to lower neighbors. The bottom star is replaced by a similarity function defined on the two sets of attributes that correspond to particular objects ($SMC$ and $J$ denotes simple matching coefficient and Jaccard similarity, respectively). The first and the second top stars are replaced by cohesion functions defined on the set of neighbouring concepts and objects, respectively. |
| $P$ (predictability, 3.2) | Defined the same way as the previous one, the only difference is the cohesion function is replaced by a predictability function. |
| CV (cue validity, 3.2) | The sum of the conditional probabilities of objects from an extent given attributes from an intent. |
| CFC (category feature collocation, 3.2) | Characterizes specificity of all attributes of a context for given set of objects. |
| CU (category utility, 3.2) | Characterizes specificity each context attribute for a given concept. |